\pdfoutput=1

\documentclass[11pt]{article}

\usepackage[preprint]{ACL}

\usepackage{times}
\usepackage{subcaption}
\usepackage{array}
\usepackage{latexsym}
\usepackage{multirow}
\usepackage{booktabs}
\usepackage{microtype}
\usepackage{multirow}
\usepackage{latexsym}
\usepackage{booktabs}
\usepackage{courier}
\usepackage{amsmath}
\usepackage{placeins}
\usepackage{tikz}
\usepackage{hyperref}
\usepackage{fontawesome5}
\usepackage[T1]{fontenc}

\usepackage[utf8]{inputenc}

\usepackage{microtype}

\usepackage{inconsolata}

\usepackage{listings}
\usepackage{xcolor}

\definecolor{codegreen}{rgb}{0,0.6,0}
\definecolor{codegray}{rgb}{0.5,0.5,0.5}
\definecolor{codepurple}{rgb}{0.58,0,0.82}
\definecolor{backcolour}{rgb}{1,1,1}

\lstdefinestyle{mystyle}{
    backgroundcolor=\color{backcolour},   
    commentstyle=\color{codegreen},
    keywordstyle=\color{magenta},
    numberstyle=\tiny\color{codegray},
    stringstyle=\color{codepurple},
    basicstyle=\ttfamily\footnotesize,
    breakatwhitespace=false,         
    breaklines=true,                 
    captionpos=b,                    
    keepspaces=true,                 
    numbers=left,                    
    numbersep=5pt,                  
    showspaces=false,                
    showstringspaces=false,
    showtabs=false,                  
    tabsize=2
}

\lstdefinelanguage{Isabelle}{
  alsoletter={<,>},
  morekeywords={
    theorem, assumes, from, then, have, shows, proof, qed, using, show,
    by, blast
  },
  sensitive=true,
  comment=[l]{(*},
  morecomment=[s]{(*}{*)},
  morestring=[b]"
}

\definecolor{darkgreen}{RGB}{0,100,0}
\definecolor{gray}{rgb}{0.5,0.5,0.5}

\lstdefinestyle{isastyle}{
    language=Isabelle,
    alsoletter={<,>},
    basicstyle=\scriptsize\ttfamily,
    keywords={theorem, assumes, from, then, have, shows, proof, qed, using, show},
    keywordstyle=\color{blue}\bfseries,
    keywords=[2]{<ATP>},
    keywordstyle=[2]{\color{red}\bfseries},
    keywords=[3]{by, blast},
    keywordstyle=[3]{\color{darkgreen}\bfseries},
    comment=[l]{(*},
    commentstyle=\color{gray}\itshape,
    morecomment=[s]{(*}{*)},
    mathescape=true,
    showstringspaces=false,
    breaklines=true,
    frame=single,
    numberstyle=\tiny,
    tabsize=2
}

\lstnewenvironment{isabellecode}{%
  \lstset{style=isastyle}%
}{}

\makeatletter
\newcommand{\printfnsymbol}[1]{%
  \textsuperscript{\@fnsymbol{#1}}%
}

\title{PEIRCE: Unifying Material and Formal Reasoning \\via LLM-Driven Neuro-Symbolic Refinement}

\author{ \textbf{Xin Quan\thanks{Equal contribution}\textsuperscript{1}, Marco Valentino\printfnsymbol{1}\textsuperscript{2}},
\\ \textbf{Danilo S. Carvalho\textsuperscript{1,3}, Dhairya Dalal\textsuperscript{4}, Andr\'e Freitas\textsuperscript{1,2,3}} \\
\textsuperscript{1}University of Manchester, United Kingdom \\ 
\textsuperscript{2}Idiap Research Institute, Switzerland \\
\textsuperscript{3}National Biomarker Centre, CRUK-MI, University of Manchester, United Kingdom \\
\textsuperscript{4}University of Galway, Ireland \\
\faGithub \textit{ }
    \url{https://github.com/neuro-symbolic-ai/peirce/}}

\begin{document}
\maketitle

\begin{abstract}
A persistent challenge in AI is the effective integration of material and formal inference -- the former concerning the plausibility and contextual relevance of arguments, while the latter focusing on their logical and structural validity. Large Language Models (LLMs), by virtue of their extensive pre-training on large textual corpora, exhibit strong capabilities in material inference. However, their reasoning often lacks formal rigour and verifiability. At the same time, LLMs’ linguistic competence positions them as a promising bridge between natural and formal languages, opening up new opportunities for combining these two modes of reasoning.
In this paper, we introduce PEIRCE, a neuro-symbolic framework designed to unify material and formal inference through an iterative conjecture–criticism process. Within this framework, LLMs play the central role of generating candidate solutions in natural and formal languages, which are then evaluated and refined via interaction with external critique models. These critiques include symbolic provers, which assess formal validity, as well as soft evaluators that measure the quality of the generated arguments along linguistic and epistemic dimensions such as plausibility, coherence, and parsimony. While PEIRCE is a general-purpose framework, we demonstrate its capabilities in the domain of natural language explanation generation -- a setting that inherently demands both material adequacy and formal correctness.
\end{abstract}

\section{Introduction}

A core challenge in Artificial Intelligence (AI) is the integration of material and formal inference \cite{mahowald2024dissociating,guo2025deepseek,cheng2025empowering, dasgupta2022language,valentino-freitas-2024-introductory,hamilton2024neuro,kambhampati2024position}. Drawing from classical distinctions in logic and philosophy of science \cite{brandom1994making,haack1978philosophy}, formal inference concerns the structural validity of arguments -- whether conclusions follow necessarily from a set of premises according to fixed syntactic rules -- while material inference is concerned with the plausibility of those arguments and their grounding in background knowledge, context, and domain-specific assumptions. Despite their complementary nature, these forms of inference are typically handled by distinct types of systems in AI: symbolic provers for formal reasoning, and statistical or neural models for material inference.

Recently, the advent of Large Language Models (LLMs) offers new opportunities for bridging these two modalities \cite{xu-etal-2024-faithful, Gandarela2024InductiveLO, NEURIPS2024_8678da90, Ranaldi2025ImprovingCR}. Their linguistic fluency and access to broad world knowledge, in fact, enable them to generate candidate solutions that approximate material reasoning. Simultaneously, emerging work has shown that LLMs can support autoformalisation, translating natural language content into structured logical forms suitable for downstream symbolic verification \cite{quan-etal-2024-verification, pan-etal-2023-logic, olausson-etal-2023-linc, jiang-etal-2024-leanreasoner, kirtania-etal-2024-logic}. This creates an opportunity for hybrid neuro-symbolic architectures that leverage the interpretive strengths of LLMs alongside the rigour of symbolic solvers.

\begin{figure*}[t]
\centering
    \includegraphics[width=\textwidth]{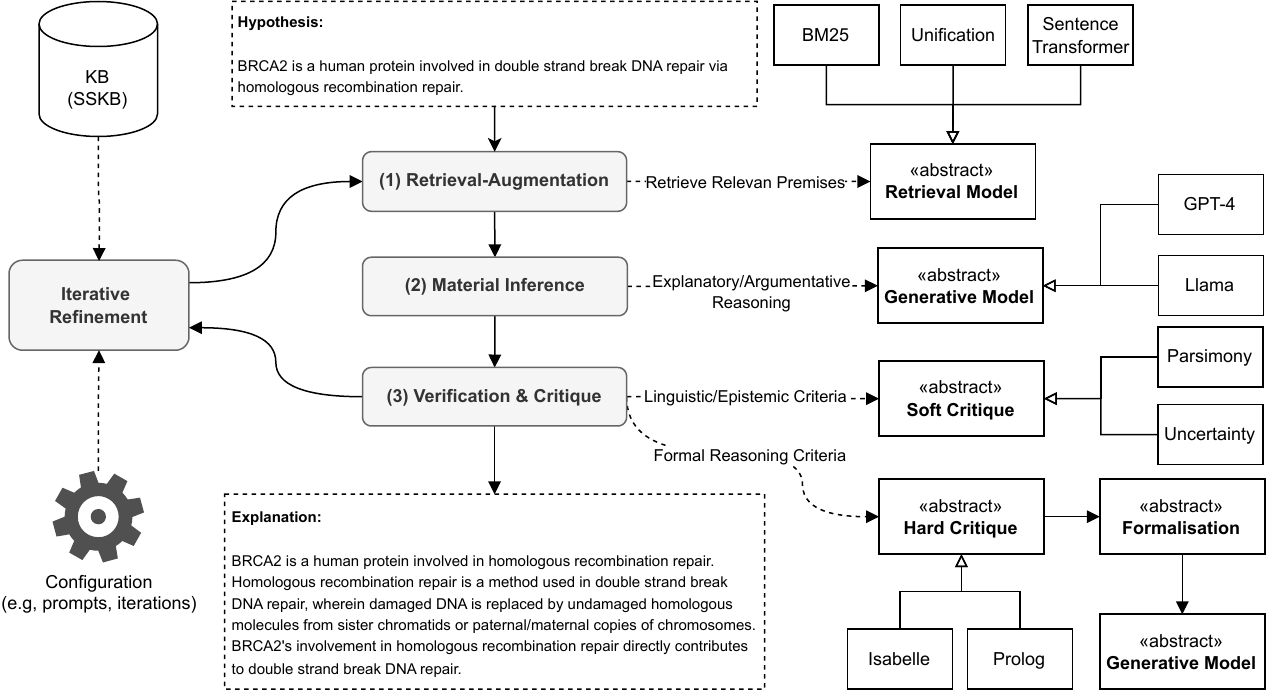}
    \caption{Overall architecture of PEIRCE. The framework provides an extensible and modular environment for unifying material and formal inference in natural language via a \emph{conjecture-criticism} process. PEIRCE supports controllability and formal error correction mechanisms for implementing a complete end-to-end iterative refinement pipeline for explanatory arguments generated by LLMs.}
    \label{fig:pipeline}
\end{figure*}

This paper presents PEIRCE, a modular and extensible framework for modelling iterative reasoning workflows that unify material and formal inference. PEIRCE implements a conjecture–criticism cycle, in which LLMs generate candidate solutions in natural and formal languages, and a suite of external critique models -- ranging from formal proof assistants to linguistic and semantic evaluators -- assessing the quality of the generated solutions according to multiple criteria, including logical validity, plausibility, coherence, and parsimony.

To demonstrate the capabilities of PEIRCE, we focus on the task of natural language explanation generation as a representative case study. Explanations constitute a particularly useful testbed for reasoning, as they must simultaneously satisfy formal and material constraints \cite{Valentino2024-VALOTN}. We evaluated the framework across several domains and datasets spanning from textual entailment \cite{NEURIPS2018_4c7a167b}, scientific question answering \cite{jansen2020worldtree, dalvi2021explaining}, and clinical hypothesis verification, showing how PEIRCE effectively enables the generation, evaluation and refinement of high-quality explanatory arguments.

\section{PEIRCE: Unifying Material and Formal Reasoning}


PEIRCE provides an extensible and modular environment for modelling and unifying \emph{material and formal reasoning} via a \emph{conjecture-criticism} cycle.
The overall architecture of PEIRCE is illustrated in Figure \ref{fig:pipeline}. The core functionality offered by the framework is the automation of an \emph{iterative refinement} pipeline for \emph{natural language inference} tasks in different domains. This pipeline is typically organised into three distinct stages implemented through the orchestration of customisable components -- i.e., (1) \emph{retrieval-augmentation}, (2) \emph{material inference}, and (3) \emph{verification and critique}. 

Given an NLI problem as input (e.g., answering a question, predicting an entailment relation, verifying a scientific claim or a hypothesis, etc.), the first stage in the process involves querying external knowledge bases (Section \ref{sec:explanation_corpora}) via retrieval models (Section \ref{sec:retrieval_models}) to select relevant premises to support reasoning. Subsequently, the retrieved knowledge can be provided in context to a generative model to generate an approximate solution in natural language (Section \ref{sec:generative_models}). The solution proposed by the generative model is then criticised by a suite of hard and soft critique models, which might use an internal formalisation stage (Section \ref{sec:critique_models}). The critiques' feedback can then be fed back to the generative model to refine the solution in the next iteration and improve its quality (Section \ref{sec:iterative_refinement}). 

PEIRCE provides abstract interfaces to instantiate and customise the iterative refinement pipeline, facilitating modularity and extensibility. 

\subsection{Data Model}
\label{sec:explanation_corpora}

PEIRCE integrates a data model interface designed for storing and retrieving knowledge from corpora of annotated premises. The data model is designed to be general, efficient, and extensible in order to cover a diverse set of knowledge bases supporting explanatory reasoning in different domains. 

A knowledge base consists of a sequence of $statements$ that can be loaded and navigated as a collection. A $statement$ is a single fact, a sentence, or a claim (e.g., ``The `(set) difference' between two sets $S$ and $T$ is written $S \setminus T$, and means...''), which may refer to concrete $entities$, and may be linked to a set of premises (other $statements$) which together constitute an explanation of why the statement holds (see Figure \ref{fig:sskb_diag}).

This recursive structure facilitates access to multiple datasets in a unified format oriented towards explanatory reasoning. It is implemented in the form of the \textit{Simple Statement Knowledge Bases} (SSKB) python package\footnote{\url{https://github.com/neuro-symbolic-ai/SSKB}}, illustrated in Figure~\ref{fig:sskb_diag}. SSKB includes loaders for a few popular NLI datasets, such as e-SNLI~\cite{NEURIPS2018_4c7a167b}, WorldTree~\cite{jansen2018worldtree}, ProofWiki~\cite{ferreira2020natural}, EntailmentBank~\cite{dalvi2021explaining}, and NLI4CT~\cite{jullien2023nli4ct,jullien-etal-2023-semeval,jullien2024semeval} and also facilitates linguistic annotations through its compatibility with the \textit{Simple Annotation Framework} (SAF)\footnote{\url{https://github.com/dscarvalho/saf}} NLP package.

\subsection{Retrieval Models}
\label{sec:retrieval_models}

In order to support the retrieval of relevant premises for reasoning from the knowledge base, PEIRCE provides an interface for implementing a suite of retrieval models, including sparse (i.e., BM25 \cite{robertson1995okapi}), dense (i.e., Sentence-Transformers \cite{reimers2019sentence}) and hybrid models specialised for explanatory inference (i.e., Unification and SCAR \cite{valentino2021unification,valentino2022hybrid}). The retrieval models are fully integrated with the data model to enable a dialogue with external corpora. 
Moreover, PEIRCE supports the creation of hybrid ensembles between retrieval models, allowing for a weighted ranking function (see Appendix \ref{sec:appendix_retrieval} for a concrete example).

\subsection{Generative Models}
\label{sec:generative_models}

PEIRCE implements a suite of classes to efficiently prompt and manage the adoption of different families of LLMs. In particular, PEIRCE supports full compatibility with OpenAI\footnote{\url{https://openai.com/index/openai-api/}} and Huggingface\footnote{\url{https://huggingface.co/models}} models. Different specialised classes following the same abstract interface facilitate reusability and extensibility for prompting LLMs for iterative refinement. 
The generative models internally use a class for dynamic prompting management that allows for the runtime instantiation of specific variables. This mechanism allows for the definition of a single prompt template that can be adapted at execution time to run experiments on different NLI problems (see Appendix \ref{sec:appendix_generative} for a concrete example).

\subsection{Critique Models}
\label{sec:critique_models}

The critique models are at the core of the iterative refinement process implemented in PEIRCE, representing the mechanism adopted to identify errors, inconsistencies and to determine the quality of the solutions generated by the LLMs. To facilitate their implementation and reuse, PEIRCE provides a suite of critique models, which can be instantiated and invoked through a common interface. In particular, PEIRCE provides the possibility of implementing both hard and soft critiques \cite{kambhampati2024position,dalal-etal-2024-inference}. 

A hard critique model is responsible for verifying formal aspects of the reasoning, such as logical validity, and typically returns a discrete value (i.e., 1 or 0) that characterises the correctness of a specific aspect. Because of their formal nature, hard critique models may use an internal formalisation process to convert natural language into machine-verifiable languages (e.g., first-order logic).  
A soft critique model, on the other hand, is responsible for analysing linguistic and stylistic aspects of the generated solution (e.g., simplicity, uncertainty) and returns a normalised continuous score that quantifies the presence of a particular feature. Contrary to hard critique models, soft critiques do not typically require formalisation and operate directly on generated arguments in natural language.

A series of information can be returned within a critique model's output depending on its nature, including a quality score in the case of a soft critique or the results of a formal verification (e.g., a logical proof) in the case of a hard critique. A concrete example of implementation is available in Appendix \ref{sec:appendix_critique}.

\subsubsection{Hard Critiques} 

Following recent work on the integration of LLMs and proof assistants for the verification and refinement of explanations \cite{quan-etal-2024-verification,quan2024enhancing}, PEIRCE provides a built-in implementation of hard critique models based on Isabelle\footnote{\url{https://isabelle.in.tum.de/}} and Prolog\footnote{\url{https://www.swi-prolog.org/}}.

These models use an internal formalisation process (through LLMs) to convert the NLI problem and the generated explanatory argument into a formal theory (through axioms and theorems) and verify, using a proof assistant or a symbolic solver, whether the generated solution logically entails the problem. If this is the case, the critique models will judge the solution as logically valid and will return the proof tactics found by the solver. If a proof cannot be found, the critique models return a detailed feedback describing the steps in which the proof construction has failed, allowing for error correction in a subsequent iteration. 

The following is an example of proof tactics returned by the \texttt{IsabelleSolver} after successful verification:

\begin{lstlisting}
'proof tactics': ['Sledgehammering...', 'cvc4 found a proof...', 'cvc4: Try this: using assms explanation_1 explanation_2 by blast (1 ms)', 'vampire found a proof...', 'vampire: Found duplicate proof', 'spass found a proof...', 'spass: Found duplicate proof', 'zipperposition found a proof...', 'zipperposition: Found duplicate proof', 'Done']
\end{lstlisting}

\begin{table}[t]
\centering
\small
\begin{tabular}{@{}lcc@{}}  
\toprule
& \textbf{Science QA} & \textbf{Premise Selection} \\
\midrule
BM25 & 22.84 & 10.18\\
Unification & 30.40 & 24.45\\
BM25 + Unification & \textbf{38.72} & \textbf{27.09}\\
\bottomrule
\end{tabular}
\caption{Explanation retrieval results (i.e., MAP) for science question answering (i.e., WorldTree) and natural language premise selection (i.e., ProofWiki).}
\label{tab:explanation_retrieval}
\end{table}

\begin{figure*}[t]
\centering
\begin{subfigure}{.32\textwidth} 
  \centering
  \includegraphics[width=\linewidth]{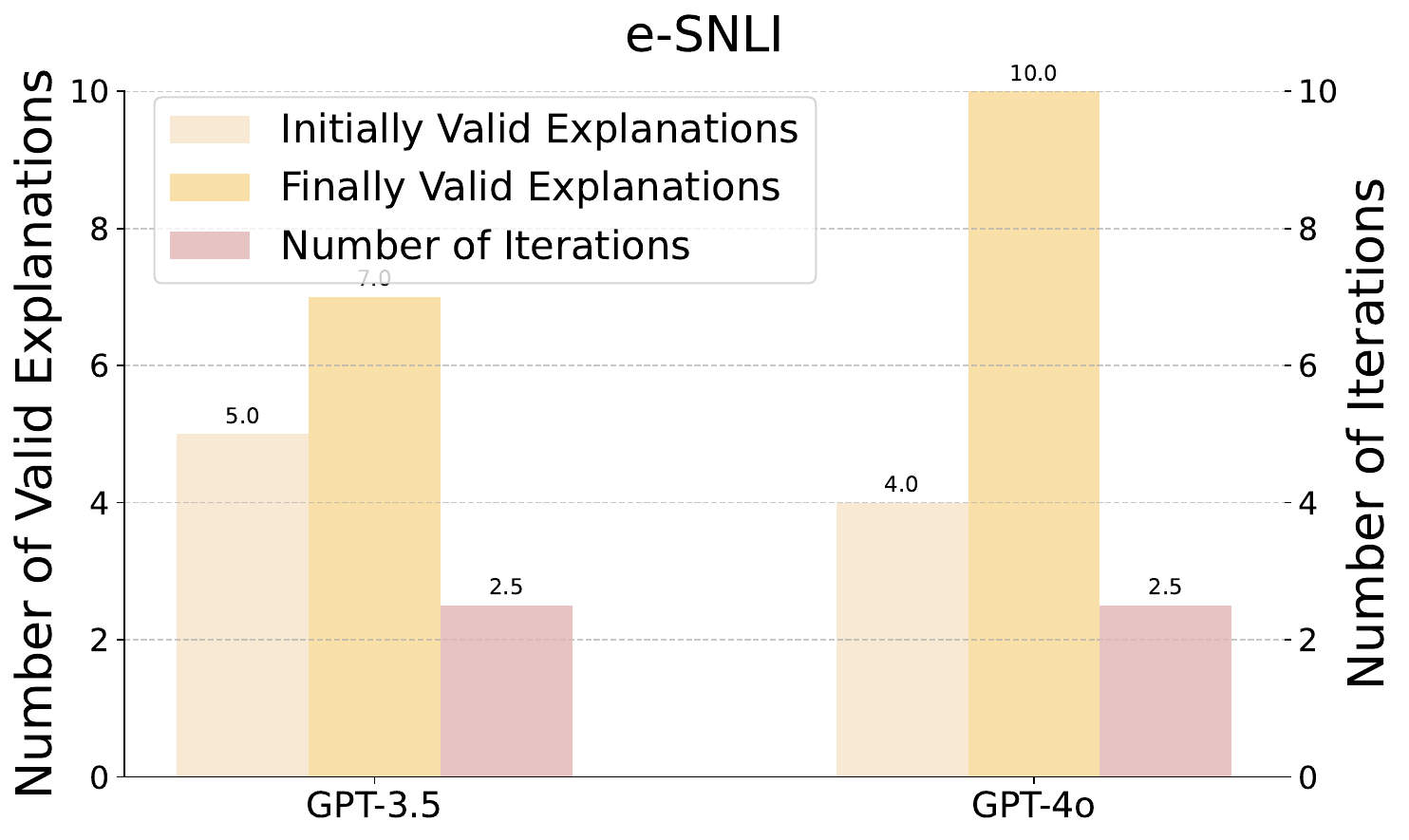}
  \caption{}
  \label{fig:sub1_esnli}
\end{subfigure}
\hfill
\begin{subfigure}{.32\textwidth} 
  \centering
  \includegraphics[width=\linewidth]{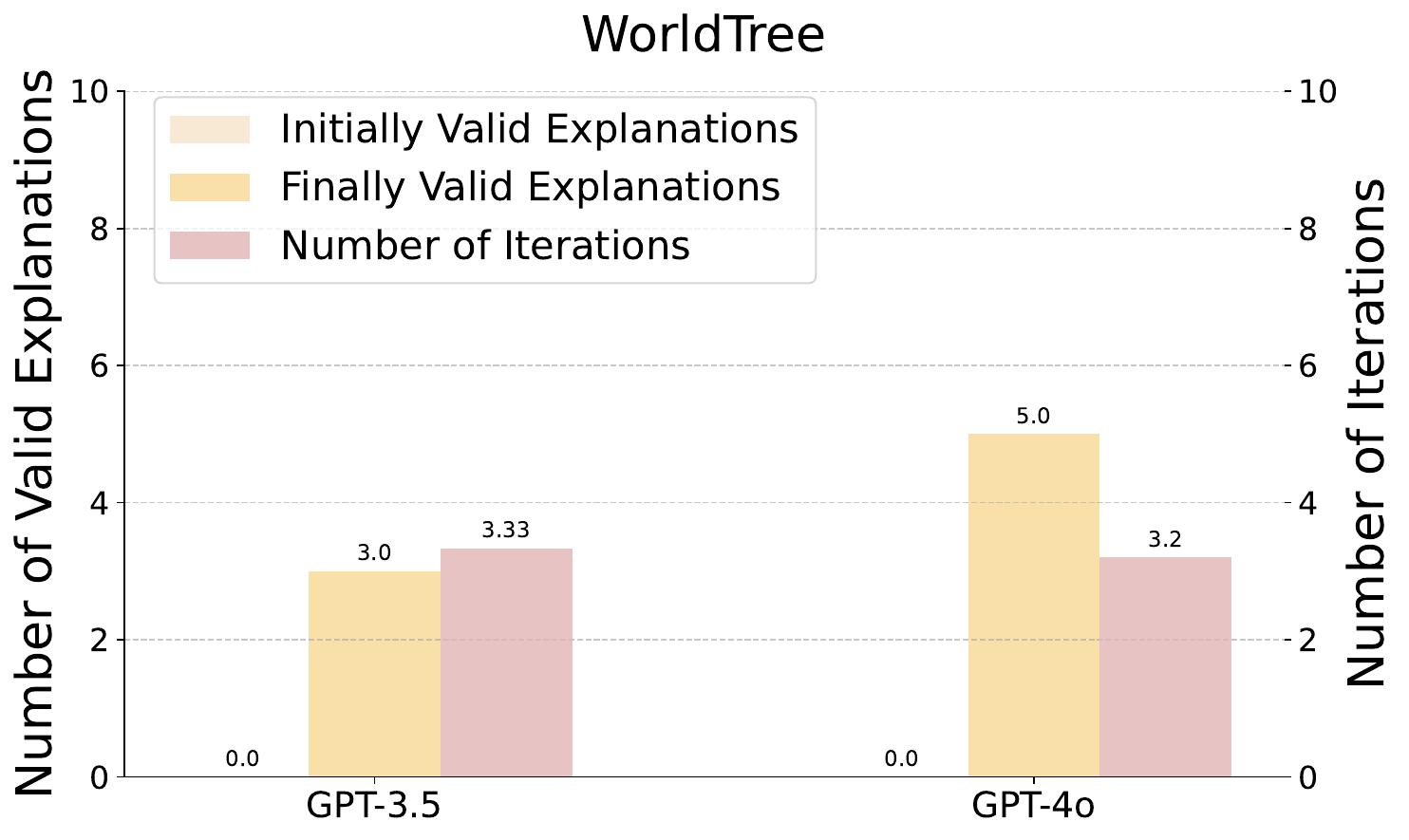}
  \caption{}
  \label{fig:sub2_worldtree}
\end{subfigure}
\hfill
\begin{subfigure}{.32\textwidth} 
  \centering
  \includegraphics[width=\linewidth]{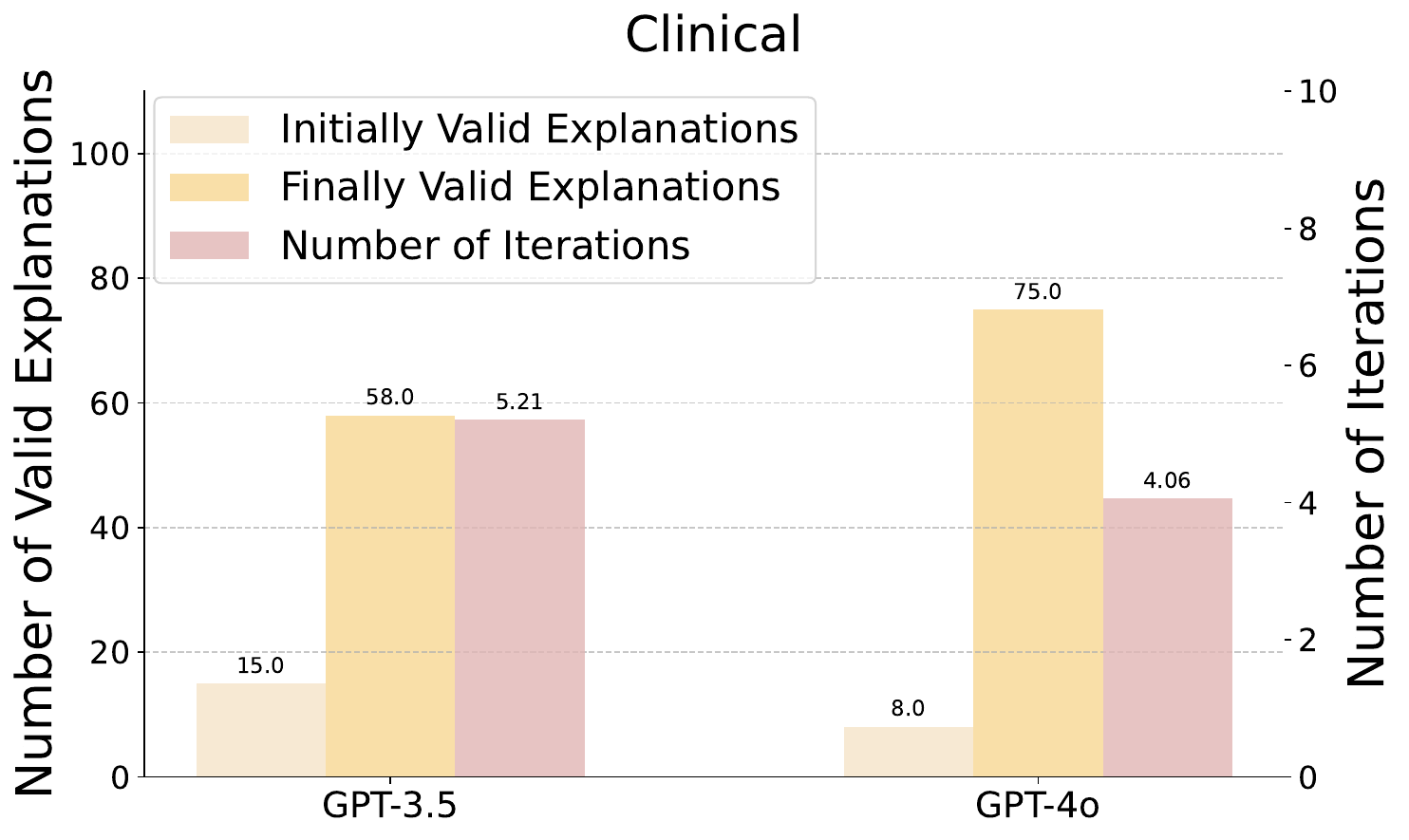}
  \caption{}
  \label{fig:sub3_clinical}
\end{subfigure}
\caption{Explanation refinement results via hard critique using GPT-4o and Isabelle (i.e., number of successfully verified explanations after a maximum of 10 iterations).}
\label{fig:results_explanation_refinement}
\end{figure*}

\begin{table*}[t]
\centering
\small
\begin{tabular}{@{}p{1cm}p{5.5cm}p{5.5cm}p{1cm}p{1cm}@{}}  
\toprule
\textbf{Dataset} & \textbf{Problem} & \textbf{Explanation} & \textbf{Iteration} & \textbf{Validity} \\
\midrule
e-SNLI & \textbf{Premise}: An infant is in a crib and crying. \newline \textbf{Hypothesis}: A baby is unhappy. & if the infant is crying, it can be assumed that they are unhappy. &\centering 0\arraybackslash & Invalid \\\\
 & & if the infant is crying, it can be assumed that they are unhappy. An infant is a type of baby. &\centering1\arraybackslash & Valid \\
\bottomrule
\end{tabular}
\caption{An example of how the explanations in e-SNLI can be refined via hard critique (i.e., GPT-4o and Isabelle).}
\label{tab:e-snli_example_1_table}
\end{table*}

\subsubsection{Soft Critiques} 
\label{sec: soft-critique}
Soft critiques are inspired by argumentation theory \cite{vanEemeren2014}
and philosophical accounts of inference to best explanation \cite{thagard1978best,lipton2017inference}. Such methods can be adopted to qualify explanatory arguments and provide comparable selection criteria to identify the best solution amongst competing hypotheses. PEIRCE provides a built-in implementation of the parsimony, coherence, and uncertainty critique models introduced by \citet{dalal-etal-2024-inference}.

\paragraph{Parsimony.} Also known as Ockam's razor, parsimony favours arguments with the fewest assumptions and premises. This soft critique model is implemented computing the \emph{concept drift}, which measures the number of new concepts and entities not present in the original NLI problem that are introduced in the generated solution.

\paragraph{Coherence} Coherence evaluates the intermediate entailment relationships between the generated premises, favouring arguments that introduce conditional clauses that are more plausible. Specifically, this critique model adopts a pre-trained textual entailment model to measure the average entailment strength (through the predicted entailment score) over generated if-then clauses in an explanatory argument. 

\paragraph{Uncertainty} Uncertainty evaluates the plausibility of a generated argument via explicit linguistic signalling expressions. In particular, this critique models analyses hedging words such as \textit{probably}, \textit{might be}, and \textit{could be} that typically signal ambiguity and are often used when the truth condition of a statement is unknown or probabilistic. This critique model adopts a fine-tuned model which analyses hedging language to establish the degree of uncertainty in the generated statements \cite{pei2021measuring}.

\subsection{Iterative Refinement}
\label{sec:iterative_refinement}

Finally, PEIRCE provides a customisable class for iterative refinement that flexibly combines the components responsible for each intermediate stage.

In particular, a class named \texttt{RefinementModel} is responsible for orchestrating retrieval models, LLMs, and critique models to perform solution refinement for a fixed number of iterations. If the critique model performs a hard critique (e.g., Isabelle), the refinement process ends when the generated argument can be formally verified (e.g., a proof is found). After the refinement, the output of the critique models, as well as the solution produced at each iteration step, will be returned. An example of implementation can be found in Appendix \ref{sec:appendix_refinement}.

\begin{figure*}[t]
\centering
\begin{subfigure}{.35\textwidth} 
  \centering
  \includegraphics[width=\linewidth]{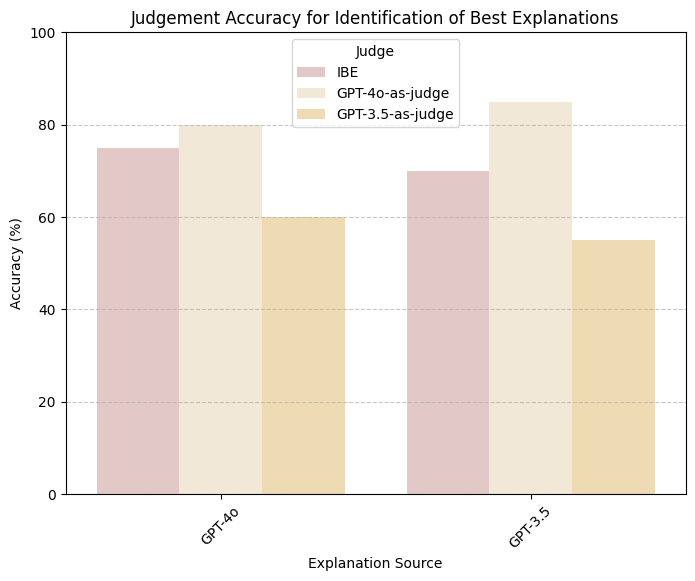}
  \caption{}
  \label{fig:soft-critique-judgement}
\end{subfigure}
\hfill
\begin{subfigure}{.64\textwidth} 
  \centering
  \includegraphics[width=\linewidth]{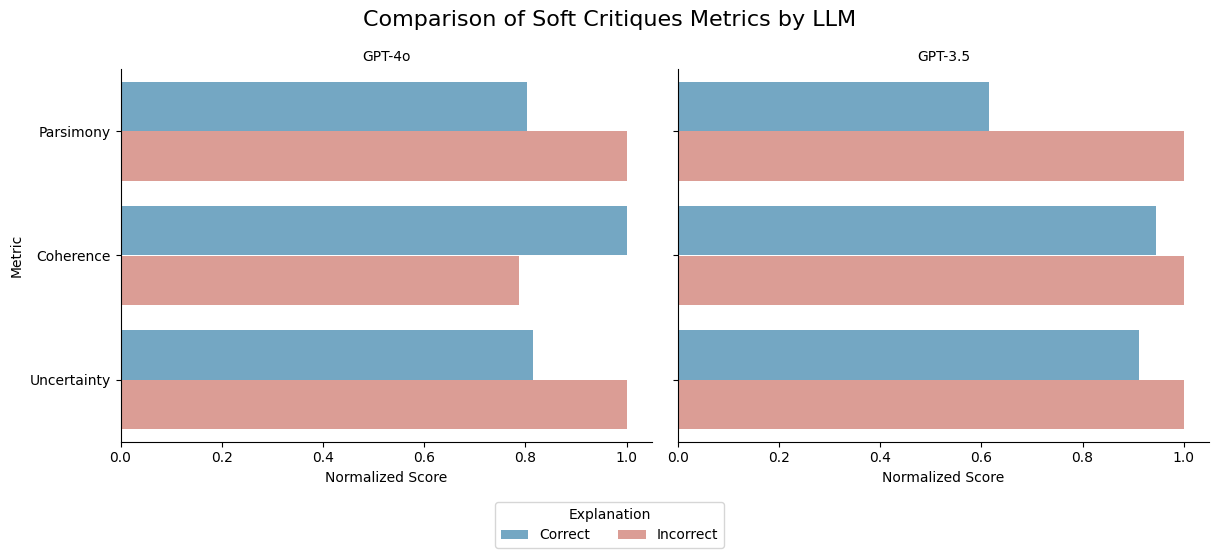}
  \caption{}
  \label{fig:soft-crtique-comp}
\end{subfigure}
\caption{Accuracy in identifying the explanation associated with the correct answer via different soft critique models (i.e., parsimony, coherence and uncertainty in IBE vs. GPT-as-judge).}
\label{fig:soft-critique-results}
\end{figure*}

\begin{table*}[]
\resizebox{\textwidth}{!}{%
\begin{tabular}{@{}lllll@{}}
\toprule
\multicolumn{5}{c}{\begin{tabular}[c]{@{}c@{}}\textbf{The cat chased the bird? What was the effect?}\\ A) The bird flew away. B) The bird caught a worm\end{tabular}}         \\ \midrule
 & \multicolumn{1}{c}{\textbf{Explanation 1}} & \multicolumn{1}{c}{\textbf{Scores}} & \multicolumn{1}{c}{\textbf{Explanation 2}} & \multicolumn{1}{c}{\textbf{Scores}} \\ \midrule
GPT-4o &
  \begin{tabular}[c]{@{}l@{}} If a cat chases a bird, then the bird may perceive a threat.\\ If the bird perceives a threat, then it is likely to take \\ evasive action to escape.\\ If the bird takes evasive action to escape, then it may \\ fly away from the area.\\ If the bird flies away from the area, then it will no longer \\ be in the vicinity of the cat.\\ Therefore, since the cat chased the bird, the bird perceived \\ a threat and took evasive action by flying away to escape \\ from the cat.\end{tabular} &
  \begin{tabular}[c]{@{}l@{}}Coherence: 0.25\\ Parsimony: 6\\ Uncertainty: 1.03\end{tabular} &
  \begin{tabular}[c]{@{}l@{}} If a cat chases a bird, then the bird may become \\ startled or distracted.\\ If the bird is startled or distracted, then it may stop \\ focusing on its current activity, such as foraging for food.\\ If the bird stops focusing on foraging, then it may \\ inadvertently come across other food sources, such as worms.\\ If the bird comes across a worm, then it may catch the worm \\ as a food source.\\ Therefore, since the cat chased the bird, causing it to become \\ startled and stop foraging, the bird may have inadvertently caught \\ a worm as a result of this distraction.\end{tabular} &
  \begin{tabular}[c]{@{}l@{}}Coherence: 0.09 \\ Parsimony: 9\\ Uncertainty: 2.33\end{tabular} \\ \midrule
GPT-3.5 &
  \begin{tabular}[c]{@{}l@{}} If a cat chases a bird, then the bird may feel threatened.\\ If the bird feels threatened, then it may try to escape.\\ If the bird tries to escape, then it may fly away.\\ Therefore, since the cat chased the bird, causing it to \\ feel threatened, the bird likely flew away as a natural \\ response to escape from the perceived danger.\end{tabular} &
  \begin{tabular}[c]{@{}l@{}}Coherence: 0.06\\ Parsimony: 1\\ Uncertainty: 1.39\end{tabular} &
  \begin{tabular}[c]{@{}l@{}} If the cat chased the bird, then the bird may have been alerted \\ and flown away.\\ If the bird flew away, then it may have landed on the ground to \\ search for food.\\ If the bird landed on the ground to search for food, then it may \\ have caught a worm.\\ Therefore, if the cat chased the bird, causing it to fly away and land \\ on the ground to search for food, it is plausible that the bird caught a \\ worm during its search.\end{tabular} &
  \begin{tabular}[c]{@{}l@{}}Coherence: -0.05\\ Parsimony: 2\\ Uncertainty: 1.65\end{tabular} \\ \bottomrule
\end{tabular}%
}
\caption{An example of evaluating competing explanations via IBE using different soft critiques.}
\label{tab:soft-critique-ex}
\end{table*}

\section{Empirical Evaluation}

We performed experiments to showcase PEIRCE's applicability to explanation-based NLI problems in different domains. In particular, we adopt PEIRCE to reproduce relevant models for natural language explanation generation, focusing on explanation retrieval, neuro-symbolic refinement of explanations for NLI, and inference to the best explanation with LLMs.

\subsection{Explanation Retrieval}

For explanation retrieval, we measure the performance of BM25 \cite{robertson2009probabilistic}, the Unification-based retrieval model \cite{valentino2021unification,valentino2022hybrid}, and an ensemble between the two on Science Question Answering (QA) and Natural Language Premise Selection. 
To this end, we measure the Mean Average Precision (MAP) of the retrieved explanatory premises on 50 randomly selected examples from the WorldTree corpus (for Science QA) \cite{jansen2018worldtree,jansen2020worldtree,jansen2021textgraphs} and ProofWiki (for Premise Selection) \cite{ferreira2020natural,valentino2022textgraphs}.
The results, reported in Table \ref{tab:explanation_retrieval}, confirm the impact of the Unification-based retrieval model reported in previous work \cite{valentino-etal-2021-natural,valentino-etal-2022-case,valentino2022hybrid}, also demonstrating the benefit of performing an ensemble between the models.

\subsection{Iterative Refinement via Hard Critique}

Using the built-in implementation of the refinement model and the hard critique based on Isabelle, we reproduced the iterative refinement pipeline introduced by \cite{quan-etal-2024-verification} on different domains (i.e., general textual entailment on e-SNLI \cite{NEURIPS2018_4c7a167b}, science questions on Worldtree \cite{jansen2018worldtree}, and clinical explanations annotated by domain experts). In particular, Figure \ref{fig:results_explanation_refinement} shows the number of natural language explanations that can be successfully verified and refined through the interaction of GPT-4o\cite{achiam2023gpt} and Isabelle \cite{nipkow2002isabelle} after a maximum of 10 iterations. Qualitative examples of the results of the refinement process are provided in Tables \ref{tab:e-snli_example_1_table} and \ref{tab:examples_appendix_table}.

\subsection{Inference to the Best Explanation via Soft Critique}

Finally, we demonstrate how soft critique models can be used to perform inference to the best explanation with LLMs \cite{dalal-etal-2024-inference}. Here, we consider the task of cause and effect prediction in a multiple-choice setting, where given a question and two competing candidates, the LLM must decide which is the most plausible answer. To this end, 20 causal questions were sourced from COPA \cite{gordon-etal-2012-semeval}. GPT-4o and GPT-3.5 are then tasked with generating causal explanations for each candidate, which are then evaluated using the soft-critique criteria (Section \ref{sec: soft-critique}). The best explanation is selected via a majority vote through the soft-critique scores (see example in Table \ref{tab:soft-critique-ex}). For comparison, LLM-as-judge baselines are provided in Figure \ref{fig:soft-critique-judgement}, with the results of the soft critique metrics reported provided in Figure \ref{fig:soft-crtique-comp}. 

\subsection{Related Work}

Neuro-symbolic reasoning models integrate neural networks with symbolic solvers to provide a reliable and verifiable reasoning process for complex downstream tasks (e.g., multi-hop reasoning, scientific question-answering) involving large datasets \cite{minervini2020learningreasoningstrategiesendtoend,Kalyanpur2020BraidWS,shi-etal-2021-neural,wang-pan-2022-deep,ijcai2024p0399}.

Several studies have proposed differentiable solvers that enhance both the robustness of rule-based models and the interpretability of neural models \cite{rocktaschel2017end, NEURIPS2018_dc5d637e, weber-etal-2019-nlprolog, thayaparan-etal-2022-diff}. More recently, integrating LLMs with logical reasoners has demonstrated significant effectiveness on natural language datasets \cite{gandarela2025inductivelearninglogicaltheories, dalal-etal-2024-inference,lyu2023faithful}. 

Research efforts have applied LLMs for autoformalisation, converting natural language into first-order logic forms, and subsequently employing symbolic provers on logical reasoning datasets \citep{pan-etal-2023-logic, olausson-etal-2023-linc, jiang-etal-2024-leanreasoner}. \citet{quan-etal-2024-verification} integrated LLMs with external theorem provers for open-world natural language inference tasks to verify and refine natural language explanations. 

Our research incorporates soft and hard critique models that uses various symbolic solvers and LLMs to evaluate logical and linguistic features, ensuring delivering logically valid, sound, and consistent explanations.

\subsection{Conclusion \& Future Work}

This paper introduced PEIRCE, a framework that provides an extensible and modular environment for unifying material and formal inference in natural language via a \emph{conjecture-criticism} process. PEIRCE supports controllability and formal error correction mechanisms for implementing a complete end-to-end iterative refinement pipeline for explanatory arguments generated by LLMs. We hope the release of PEIRCE will facilitate new research on neuro-symbolic applications driven by LLMs. In future work, we plan to extend the suite of ready-to-use knowledge resources and critique models in the framework as well as integrate PEIRCE with a supervised fine-tuning and reinforcement learning pipeline to leverage the feedback generated by the critique models and the refined solution for training.

\bibliography{anthology,custom}

\begin{thebibliography}{56}
\providecommand{\natexlab}[1]{#1}

\bibitem[{Achiam et~al.(2023)Achiam, Adler, Agarwal, Ahmad, Akkaya, Aleman, Almeida, Altenschmidt, Altman, Anadkat et~al.}]{achiam2023gpt}
Josh Achiam, Steven Adler, Sandhini Agarwal, Lama Ahmad, Ilge Akkaya, Florencia~Leoni Aleman, Diogo Almeida, Janko Altenschmidt, Sam Altman, Shyamal Anadkat, and 1 others. 2023.
\newblock Gpt-4 technical report.
\newblock \emph{arXiv preprint arXiv:2303.08774}.

\bibitem[{Brandom(1994)}]{brandom1994making}
Robert Brandom. 1994.
\newblock \emph{Making it explicit: Reasoning, representing, and discursive commitment}.
\newblock Harvard university press.

\bibitem[{Camburu et~al.(2018)Camburu, Rockt\"{a}schel, Lukasiewicz, and Blunsom}]{NEURIPS2018_4c7a167b}
Oana-Maria Camburu, Tim Rockt\"{a}schel, Thomas Lukasiewicz, and Phil Blunsom. 2018.
\newblock \href {https://proceedings.neurips.cc/paper_files/paper/2018/file/4c7a167bb329bd92580a99ce422d6fa6-Paper.pdf} {e-snli: Natural language inference with natural language explanations}.
\newblock In \emph{Advances in Neural Information Processing Systems}, volume~31. Curran Associates, Inc.

\bibitem[{Cheng et~al.(2025)Cheng, Li, Liu, van Rooij, Zhang, and Lin}]{cheng2025empowering}
Fengxiang Cheng, Haoxuan Li, Fenrong Liu, Robert van Rooij, Kun Zhang, and Zhouchen Lin. 2025.
\newblock Empowering llms with logical reasoning: A comprehensive survey.
\newblock \emph{arXiv preprint arXiv:2502.15652}.

\bibitem[{Dalal et~al.(2024)Dalal, Valentino, Freitas, and Buitelaar}]{dalal-etal-2024-inference}
Dhairya Dalal, Marco Valentino, Andre Freitas, and Paul Buitelaar. 2024.
\newblock \href {https://doi.org/10.18653/v1/2024.acl-long.14} {Inference to the best explanation in large language models}.
\newblock In \emph{Proceedings of the 62nd Annual Meeting of the Association for Computational Linguistics (Volume 1: Long Papers)}, pages 217--235, Bangkok, Thailand. Association for Computational Linguistics.

\bibitem[{Dalvi et~al.(2021)Dalvi, Jansen, Tafjord, Xie, Smith, Pipatanangkura, and Clark}]{dalvi2021explaining}
Bhavana Dalvi, Peter Jansen, Oyvind Tafjord, Zhengnan Xie, Hannah Smith, Leighanna Pipatanangkura, and Peter Clark. 2021.
\newblock Explaining answers with entailment trees.
\newblock In \emph{Proceedings of the 2021 Conference on Empirical Methods in Natural Language Processing}, pages 7358--7370.

\bibitem[{Dasgupta et~al.(2022)Dasgupta, Lampinen, Chan, Sheahan, Creswell, Kumaran, McClelland, and Hill}]{dasgupta2022language}
Ishita Dasgupta, Andrew~K Lampinen, Stephanie~CY Chan, Hannah~R Sheahan, Antonia Creswell, Dharshan Kumaran, James~L McClelland, and Felix Hill. 2022.
\newblock Language models show human-like content effects on reasoning tasks.
\newblock \emph{arXiv preprint arXiv:2207.07051}.

\bibitem[{Ferreira and Freitas(2020)}]{ferreira2020natural}
Deborah Ferreira and Andr{\'e} Freitas. 2020.
\newblock \href {https://aclanthology.org/2020.lrec-1.266} {Natural language premise selection: Finding supporting statements for mathematical text}.
\newblock In \emph{Proceedings of the 12th Language Resources and Evaluation Conference}, pages 2175--2182, Marseille, France. European Language Resources Association.

\bibitem[{Gandarela et~al.(2024)Gandarela, Carvalho, and Freitas}]{Gandarela2024InductiveLO}
Joao~Pedro Gandarela, Danilo~S. Carvalho, and Andr'e Freitas. 2024.
\newblock \href {https://api.semanticscholar.org/CorpusID:272310256} {Inductive learning of logical theories with llms: A complexity-graded analysis}.
\newblock \emph{ArXiv}, abs/2408.16779.

\bibitem[{Gandarela et~al.(2025)Gandarela, Carvalho, and Freitas}]{gandarela2025inductivelearninglogicaltheories}
João~Pedro Gandarela, Danilo~S. Carvalho, and André Freitas. 2025.
\newblock \href {https://arxiv.org/abs/2408.16779} {Inductive learning of logical theories with llms: An expressivity-graded analysis}.
\newblock \emph{Preprint}, arXiv:2408.16779.

\bibitem[{Gordon et~al.(2012)Gordon, Kozareva, and Roemmele}]{gordon-etal-2012-semeval}
Andrew Gordon, Zornitsa Kozareva, and Melissa Roemmele. 2012.
\newblock \href {https://aclanthology.org/S12-1052} {{S}em{E}val-2012 task 7: Choice of plausible alternatives: An evaluation of commonsense causal reasoning}.
\newblock In \emph{*{SEM} 2012: The First Joint Conference on Lexical and Computational Semantics {--} Volume 1: Proceedings of the main conference and the shared task, and Volume 2: Proceedings of the Sixth International Workshop on Semantic Evaluation ({S}em{E}val 2012)}, pages 394--398, Montr{\'e}al, Canada. Association for Computational Linguistics.

\bibitem[{Guo et~al.(2025)Guo, Yang, Zhang, Song, Zhang, Xu, Zhu, Ma, Wang, Bi et~al.}]{guo2025deepseek}
Daya Guo, Dejian Yang, Haowei Zhang, Junxiao Song, Ruoyu Zhang, Runxin Xu, Qihao Zhu, Shirong Ma, Peiyi Wang, Xiao Bi, and 1 others. 2025.
\newblock Deepseek-r1: Incentivizing reasoning capability in llms via reinforcement learning.
\newblock \emph{arXiv preprint arXiv:2501.12948}.

\bibitem[{Haack(1978)}]{haack1978philosophy}
Susan Haack. 1978.
\newblock \emph{Philosophy of Logics}.
\newblock Cambridge University Press.

\bibitem[{Hamilton et~al.(2024)Hamilton, Nayak, Bo{\v{z}}i{\'c}, and Longo}]{hamilton2024neuro}
Kyle Hamilton, Aparna Nayak, Bojan Bo{\v{z}}i{\'c}, and Luca Longo. 2024.
\newblock Is neuro-symbolic ai meeting its promises in natural language processing? a structured review.
\newblock \emph{Semantic Web}, 15(4):1265--1306.

\bibitem[{Jansen and Ustalov(2020)}]{jansen2020worldtree}
Peter Jansen and Dmitry Ustalov. 2020.
\newblock \href {https://www.aclweb.org/anthology/2020.textgraphs-1.10} {{TextGraphs~2020 Shared Task on Multi-Hop Inference for Explanation Regeneration}}.
\newblock In \emph{Proceedings of the Graph-based Methods for Natural Language Processing (TextGraphs)}, pages 85--97, Barcelona, Spain (Online). Association for Computational Linguistics.

\bibitem[{Jansen et~al.(2018)Jansen, Wainwright, Marmorstein, and Morrison}]{jansen2018worldtree}
Peter Jansen, Elizabeth Wainwright, Steven Marmorstein, and Clayton Morrison. 2018.
\newblock Worldtree: A corpus of explanation graphs for elementary science questions supporting multi-hop inference.
\newblock In \emph{Proceedings of the Eleventh International Conference on Language Resources and Evaluation (LREC 2018)}.

\bibitem[{Jiang et~al.(2024)Jiang, Fonseca, and Cohen}]{jiang-etal-2024-leanreasoner}
Dongwei Jiang, Marcio Fonseca, and Shay Cohen. 2024.
\newblock \href {https://doi.org/10.18653/v1/2024.naacl-long.416} {{L}ean{R}easoner: Boosting complex logical reasoning with lean}.
\newblock In \emph{Proceedings of the 2024 Conference of the North American Chapter of the Association for Computational Linguistics: Human Language Technologies (Volume 1: Long Papers)}, pages 7497--7510, Mexico City, Mexico. Association for Computational Linguistics.

\bibitem[{Jullien et~al.(2024)Jullien, Valentino, and Freitas}]{jullien2024semeval}
Mael Jullien, Marco Valentino, and Andr{\'e} Freitas. 2024.
\newblock \href {https://doi.org/10.18653/v1/2024.semeval-1.271} {{S}em{E}val-2024 task 2: Safe biomedical natural language inference for clinical trials}.
\newblock In \emph{Proceedings of the 18th International Workshop on Semantic Evaluation (SemEval-2024)}, pages 1947--1962, Mexico City, Mexico. Association for Computational Linguistics.

\bibitem[{Jullien et~al.(2023{\natexlab{a}})Jullien, Valentino, Frost, O{'}Regan, Landers, and Freitas}]{jullien2023nli4ct}
Mael Jullien, Marco Valentino, Hannah Frost, Paul O{'}Regan, D{\'o}nal Landers, and Andre Freitas. 2023{\natexlab{a}}.
\newblock \href {https://doi.org/10.18653/v1/2023.emnlp-main.1041} {{NLI}4{CT}: Multi-evidence natural language inference for clinical trial reports}.
\newblock In \emph{Proceedings of the 2023 Conference on Empirical Methods in Natural Language Processing}, pages 16745--16764, Singapore. Association for Computational Linguistics.

\bibitem[{Jullien et~al.(2023{\natexlab{b}})Jullien, Valentino, Frost, O{'}regan, Landers, and Freitas}]{jullien-etal-2023-semeval}
Ma{\"e}l Jullien, Marco Valentino, Hannah Frost, Paul O{'}regan, Donal Landers, and Andr{\'e} Freitas. 2023{\natexlab{b}}.
\newblock \href {https://doi.org/10.18653/v1/2023.semeval-1.307} {{S}em{E}val-2023 task 7: Multi-evidence natural language inference for clinical trial data}.
\newblock In \emph{Proceedings of the 17th International Workshop on Semantic Evaluation (SemEval-2023)}, pages 2216--2226, Toronto, Canada. Association for Computational Linguistics.

\bibitem[{Kalyanpur et~al.(2020)Kalyanpur, Breloff, and Ferrucci}]{Kalyanpur2020BraidWS}
Aditya Kalyanpur, Tom Breloff, and David~A. Ferrucci. 2020.
\newblock \href {https://api.semanticscholar.org/CorpusID:227247865} {Braid: Weaving symbolic and neural knowledge into coherent logical explanations}.
\newblock In \emph{AAAI Conference on Artificial Intelligence}.

\bibitem[{Kambhampati et~al.(2024)Kambhampati, Valmeekam, Guan, Verma, Stechly, Bhambri, Saldyt, and Murthy}]{kambhampati2024position}
Subbarao Kambhampati, Karthik Valmeekam, Lin Guan, Mudit Verma, Kaya Stechly, Siddhant Bhambri, Lucas~Paul Saldyt, and Anil~B Murthy. 2024.
\newblock Position: Llms can’t plan, but can help planning in llm-modulo frameworks.
\newblock In \emph{Forty-first International Conference on Machine Learning}.

\bibitem[{Kirtania et~al.(2024)Kirtania, Gupta, and Radhakrishna}]{kirtania-etal-2024-logic}
Shashank Kirtania, Priyanshu Gupta, and Arjun Radhakrishna. 2024.
\newblock \href {https://aclanthology.org/2024.nlrse-1.6/} {{LOGIC}-{LM}++: Multi-step refinement for symbolic formulations}.
\newblock In \emph{Proceedings of the 2nd Workshop on Natural Language Reasoning and Structured Explanations (@ACL 2024)}, pages 56--63, Bangkok, Thailand. Association for Computational Linguistics.

\bibitem[{Lipton(2017)}]{lipton2017inference}
Peter Lipton. 2017.
\newblock Inference to the best explanation.
\newblock \emph{A Companion to the Philosophy of Science}, pages 184--193.

\bibitem[{Lyu et~al.(2023)Lyu, Havaldar, Stein, Zhang, Rao, Wong, Apidianaki, and Callison-Burch}]{lyu2023faithful}
Qing Lyu, Shreya Havaldar, Adam Stein, Li~Zhang, Delip Rao, Eric Wong, Marianna Apidianaki, and Chris Callison-Burch. 2023.
\newblock Faithful chain-of-thought reasoning.
\newblock In \emph{The 13th International Joint Conference on Natural Language Processing and the 3rd Conference of the Asia-Pacific Chapter of the Association for Computational Linguistics (IJCNLP-AACL 2023)}.

\bibitem[{Mahowald et~al.(2024)Mahowald, Ivanova, Blank, Kanwisher, Tenenbaum, and Fedorenko}]{mahowald2024dissociating}
Kyle Mahowald, Anna~A Ivanova, Idan~A Blank, Nancy Kanwisher, Joshua~B Tenenbaum, and Evelina Fedorenko. 2024.
\newblock Dissociating language and thought in large language models.
\newblock \emph{Trends in cognitive sciences}.

\bibitem[{Manhaeve et~al.(2018)Manhaeve, Dumancic, Kimmig, Demeester, and De~Raedt}]{NEURIPS2018_dc5d637e}
Robin Manhaeve, Sebastijan Dumancic, Angelika Kimmig, Thomas Demeester, and Luc De~Raedt. 2018.
\newblock \href {https://proceedings.neurips.cc/paper_files/paper/2018/file/dc5d637ed5e62c36ecb73b654b05ba2a-Paper.pdf} {Deepproblog: Neural probabilistic logic programming}.
\newblock In \emph{Advances in Neural Information Processing Systems}, volume~31. Curran Associates, Inc.

\bibitem[{Minervini et~al.(2020)Minervini, Riedel, Stenetorp, Grefenstette, and Rocktäschel}]{minervini2020learningreasoningstrategiesendtoend}
Pasquale Minervini, Sebastian Riedel, Pontus Stenetorp, Edward Grefenstette, and Tim Rocktäschel. 2020.
\newblock \href {https://arxiv.org/abs/2007.06477} {Learning reasoning strategies in end-to-end differentiable proving}.
\newblock \emph{Preprint}, arXiv:2007.06477.

\bibitem[{Morishita et~al.(2024)Morishita, Morio, Yamaguchi, and Sogawa}]{NEURIPS2024_8678da90}
Terufumi Morishita, Gaku Morio, Atsuki Yamaguchi, and Yasuhiro Sogawa. 2024.
\newblock \href {https://proceedings.neurips.cc/paper_files/paper/2024/file/8678da90126aa58326b2fc0254b33a8c-Paper-Conference.pdf} {Enhancing reasoning capabilities of llms via principled synthetic logic corpus}.
\newblock In \emph{Advances in Neural Information Processing Systems}, volume~37, pages 73572--73604. Curran Associates, Inc.

\bibitem[{Nipkow et~al.(2002)Nipkow, Wenzel, and Paulson}]{nipkow2002isabelle}
Tobias Nipkow, Markus Wenzel, and Lawrence~C Paulson. 2002.
\newblock \emph{Isabelle/HOL: a proof assistant for higher-order logic}.
\newblock Springer.

\bibitem[{Olausson et~al.(2023)Olausson, Gu, Lipkin, Zhang, Solar-Lezama, Tenenbaum, and Levy}]{olausson-etal-2023-linc}
Theo Olausson, Alex Gu, Ben Lipkin, Cedegao Zhang, Armando Solar-Lezama, Joshua Tenenbaum, and Roger Levy. 2023.
\newblock \href {https://doi.org/10.18653/v1/2023.emnlp-main.313} {{LINC}: A neurosymbolic approach for logical reasoning by combining language models with first-order logic provers}.
\newblock In \emph{Proceedings of the 2023 Conference on Empirical Methods in Natural Language Processing}, pages 5153--5176, Singapore. Association for Computational Linguistics.

\bibitem[{Pan et~al.(2023)Pan, Albalak, Wang, and Wang}]{pan-etal-2023-logic}
Liangming Pan, Alon Albalak, Xinyi Wang, and William Wang. 2023.
\newblock \href {https://doi.org/10.18653/v1/2023.findings-emnlp.248} {Logic-{LM}: Empowering large language models with symbolic solvers for faithful logical reasoning}.
\newblock In \emph{Findings of the Association for Computational Linguistics: EMNLP 2023}, pages 3806--3824, Singapore. Association for Computational Linguistics.

\bibitem[{Pei and Jurgens(2021)}]{pei2021measuring}
Jiaxin Pei and David Jurgens. 2021.
\newblock Measuring sentence-level and aspect-level (un) certainty in science communications.
\newblock In \emph{Proceedings of the 2021 Conference on Empirical Methods in Natural Language Processing}.

\bibitem[{Quan et~al.(2024{\natexlab{a}})Quan, Valentino, Dennis, and Freitas}]{quan2024enhancing}
Xin Quan, Marco Valentino, Louise Dennis, and Andre Freitas. 2024{\natexlab{a}}.
\newblock \href {https://aclanthology.org/2024.eacl-long.1/} {Enhancing ethical explanations of large language models through iterative symbolic refinement}.
\newblock In \emph{Proceedings of the 18th Conference of the European Chapter of the Association for Computational Linguistics (Volume 1: Long Papers)}, pages 1--22, St. Julian{'}s, Malta. Association for Computational Linguistics.

\bibitem[{Quan et~al.(2024{\natexlab{b}})Quan, Valentino, Dennis, and Freitas}]{quan-etal-2024-verification}
Xin Quan, Marco Valentino, Louise~A. Dennis, and Andre Freitas. 2024{\natexlab{b}}.
\newblock \href {https://doi.org/10.18653/v1/2024.emnlp-main.172} {Verification and refinement of natural language explanations through {LLM}-symbolic theorem proving}.
\newblock In \emph{Proceedings of the 2024 Conference on Empirical Methods in Natural Language Processing}, pages 2933--2958, Miami, Florida, USA. Association for Computational Linguistics.

\bibitem[{Ranaldi et~al.(2025)Ranaldi, Valentino, Polonsky, and Freitas}]{Ranaldi2025ImprovingCR}
Leonardo Ranaldi, Marco Valentino, Alexander Polonsky, and Andr{\'e} Freitas. 2025.
\newblock \href {https://api.semanticscholar.org/CorpusID:276421915} {Improving chain-of-thought reasoning via quasi-symbolic abstractions}.
\newblock \emph{ArXiv}, abs/2502.12616.

\bibitem[{Reimers and Gurevych(2019)}]{reimers2019sentence}
Nils Reimers and Iryna Gurevych. 2019.
\newblock Sentence-bert: Sentence embeddings using siamese bert-networks.
\newblock \emph{arXiv preprint arXiv:1908.10084}.

\bibitem[{Robertson et~al.(2009)Robertson, Zaragoza et~al.}]{robertson2009probabilistic}
Stephen Robertson, Hugo Zaragoza, and 1 others. 2009.
\newblock The probabilistic relevance framework: Bm25 and beyond.
\newblock \emph{Foundations and Trends{\textregistered} in Information Retrieval}, 3(4):333--389.

\bibitem[{Robertson et~al.(1995)Robertson, Walker, Jones, Hancock-Beaulieu, Gatford et~al.}]{robertson1995okapi}
Stephen~E Robertson, Steve Walker, Susan Jones, Micheline~M Hancock-Beaulieu, Mike Gatford, and 1 others. 1995.
\newblock Okapi at trec-3.
\newblock \emph{Nist Special Publication Sp}, 109:109.

\bibitem[{Rockt{\"{a}}schel and Riedel(2017)}]{rocktaschel2017end}
Tim Rockt{\"{a}}schel and Sebastian Riedel. 2017.
\newblock \href {http://papers.nips.cc/paper/6969-end-to-end-differentiable-proving} {End-to-end differentiable proving}.
\newblock In \emph{Advances in Neural Information Processing Systems 30: Annual Conference on Neural Information Processing Systems 2017, 4-9 December 2017, Long Beach, CA, {USA}}, pages 3791--3803.

\bibitem[{Shi et~al.(2021)Shi, Ding, Du, Liu, and Qin}]{shi-etal-2021-neural}
Jihao Shi, Xiao Ding, Li~Du, Ting Liu, and Bing Qin. 2021.
\newblock \href {https://doi.org/10.18653/v1/2021.emnlp-main.298} {Neural natural logic inference for interpretable question answering}.
\newblock In \emph{Proceedings of the 2021 Conference on Empirical Methods in Natural Language Processing}, pages 3673--3684, Online and Punta Cana, Dominican Republic. Association for Computational Linguistics.

\bibitem[{Thagard(1978)}]{thagard1978best}
Paul~R Thagard. 1978.
\newblock The best explanation: Criteria for theory choice.
\newblock \emph{The journal of philosophy}, 75(2):76--92.

\bibitem[{Thayaparan et~al.(2022)Thayaparan, Valentino, Ferreira, Rozanova, and Freitas}]{thayaparan-etal-2022-diff}
Mokanarangan Thayaparan, Marco Valentino, Deborah Ferreira, Julia Rozanova, and Andr{\'e} Freitas. 2022.
\newblock \href {https://doi.org/10.1162/tacl_a_00508} {Diff-explainer: Differentiable convex optimization for explainable multi-hop inference}.
\newblock \emph{Transactions of the Association for Computational Linguistics}, 10:1103--1119.

\bibitem[{Thayaparan et~al.(2021)Thayaparan, Valentino, Jansen, and Ustalov}]{jansen2021textgraphs}
Mokanarangan Thayaparan, Marco Valentino, Peter Jansen, and Dmitry Ustalov. 2021.
\newblock \href {https://doi.org/10.18653/v1/2021.textgraphs-1.17} {{T}ext{G}raphs 2021 shared task on multi-hop inference for explanation regeneration}.
\newblock In \emph{Proceedings of the Fifteenth Workshop on Graph-Based Methods for Natural Language Processing (TextGraphs-15)}, pages 156--165, Mexico City, Mexico. Association for Computational Linguistics.

\bibitem[{Valentino et~al.(2022{\natexlab{a}})Valentino, Ferreira, Thayaparan, Freitas, and Ustalov}]{valentino2022textgraphs}
Marco Valentino, Deborah Ferreira, Mokanarangan Thayaparan, Andr{\'e} Freitas, and Dmitry Ustalov. 2022{\natexlab{a}}.
\newblock \href {https://aclanthology.org/2022.textgraphs-1.11/} {{T}ext{G}raphs 2022 shared task on natural language premise selection}.
\newblock In \emph{Proceedings of TextGraphs-16: Graph-based Methods for Natural Language Processing}, pages 105--113, Gyeongju, Republic of Korea. Association for Computational Linguistics.

\bibitem[{Valentino and Freitas(2024{\natexlab{a}})}]{Valentino2024-VALOTN}
Marco Valentino and Andr\'e Freitas. 2024{\natexlab{a}}.
\newblock \href {https://doi.org/10.1007/s13347-024-00775-3} {On the nature of explanation: An epistemological-linguistic perspective for explanation-based natural language inference}.
\newblock \emph{Philosophy and Technology}, 37(3):1--33.

\bibitem[{Valentino and Freitas(2024{\natexlab{b}})}]{valentino-freitas-2024-introductory}
Marco Valentino and Andr{\'e} Freitas. 2024{\natexlab{b}}.
\newblock \href {https://doi.org/10.18653/v1/2024.emnlp-tutorials.4} {Reasoning with natural language explanations}.
\newblock In \emph{Proceedings of the 2024 Conference on Empirical Methods in Natural Language Processing: Tutorial Abstracts}, pages 25--31, Miami, Florida, USA. Association for Computational Linguistics.

\bibitem[{Valentino et~al.(2021{\natexlab{a}})Valentino, Pratt-Hartmann, and Freitas}]{valentino-etal-2021-natural}
Marco Valentino, Ian Pratt-Hartmann, and Andr{\'e} Freitas. 2021{\natexlab{a}}.
\newblock \href {https://aclanthology.org/2021.iwcs-1.8} {Do natural language explanations represent valid logical arguments? verifying entailment in explainable {NLI} gold standards}.
\newblock In \emph{Proceedings of the 14th International Conference on Computational Semantics (IWCS)}, pages 76--86, Groningen, The Netherlands (online). Association for Computational Linguistics.

\bibitem[{Valentino et~al.(2022{\natexlab{b}})Valentino, Thayaparan, Ferreira, and Freitas}]{valentino2022hybrid}
Marco Valentino, Mokanarangan Thayaparan, Deborah Ferreira, and André Freitas. 2022{\natexlab{b}}.
\newblock \href {https://doi.org/10.1609/aaai.v36i10.21392} {Hybrid autoregressive inference for scalable multi-hop explanation regeneration}.
\newblock In \emph{Proceedings of the AAAI Conference on Artificial Intelligence}, volume~36, pages 11403--11411.

\bibitem[{Valentino et~al.(2021{\natexlab{b}})Valentino, Thayaparan, and Freitas}]{valentino2021unification}
Marco Valentino, Mokanarangan Thayaparan, and Andr{\'e} Freitas. 2021{\natexlab{b}}.
\newblock \href {https://doi.org/10.18653/v1/2021.eacl-main.15} {Unification-based reconstruction of multi-hop explanations for science questions}.
\newblock In \emph{Proceedings of the 16th Conference of the European Chapter of the Association for Computational Linguistics: Main Volume}, pages 200--211, Online. Association for Computational Linguistics.

\bibitem[{Valentino et~al.(2022{\natexlab{c}})Valentino, Thayaparan, and Freitas}]{valentino-etal-2022-case}
Marco Valentino, Mokanarangan Thayaparan, and Andr{\'e} Freitas. 2022{\natexlab{c}}.
\newblock \href {https://aclanthology.org/2022.coling-1.134} {Case-based abductive natural language inference}.
\newblock In \emph{Proceedings of the 29th International Conference on Computational Linguistics}, pages 1556--1568, Gyeongju, Republic of Korea. International Committee on Computational Linguistics.

\bibitem[{van Eemeren et~al.(2014)van Eemeren, Garssen, Krabbe, Snoeck~Henkemans, Verheij, and Wagemans}]{vanEemeren2014}
Frans~H. van Eemeren, Bart Garssen, Erik C.~W. Krabbe, A.~Francisca Snoeck~Henkemans, Bart Verheij, and Jean H.~M. Wagemans. 2014.
\newblock \href {https://doi.org/10.1007/978-90-481-9473-5_1} {\emph{Argumentation TheoryArgumentationtheory}}, pages 1--49.
\newblock Springer Netherlands, Dordrecht.

\bibitem[{Wang and Pan(2022)}]{wang-pan-2022-deep}
Wenya Wang and Sinno Pan. 2022.
\newblock \href {https://doi.org/10.18653/v1/2022.acl-long.343} {Deep inductive logic reasoning for multi-hop reading comprehension}.
\newblock In \emph{Proceedings of the 60th Annual Meeting of the Association for Computational Linguistics (Volume 1: Long Papers)}, pages 4999--5009, Dublin, Ireland. Association for Computational Linguistics.

\bibitem[{Weber et~al.(2019)Weber, Minervini, M{\"u}nchmeyer, Leser, and Rockt{\"a}schel}]{weber-etal-2019-nlprolog}
Leon Weber, Pasquale Minervini, Jannes M{\"u}nchmeyer, Ulf Leser, and Tim Rockt{\"a}schel. 2019.
\newblock \href {https://doi.org/10.18653/v1/P19-1618} {{NLP}rolog: Reasoning with weak unification for question answering in natural language}.
\newblock In \emph{Proceedings of the 57th Annual Meeting of the Association for Computational Linguistics}, pages 6151--6161, Florence, Italy. Association for Computational Linguistics.

\bibitem[{Weir et~al.(2024)Weir, Clark, and Van~Durme}]{ijcai2024p0399}
Nathaniel Weir, Peter Clark, and Benjamin Van~Durme. 2024.
\newblock \href {https://doi.org/10.24963/ijcai.2024/399} {Nellie: A neuro-symbolic inference engine for grounded, compositional, and explainable reasoning}.
\newblock In \emph{Proceedings of the Thirty-Third International Joint Conference on Artificial Intelligence, {IJCAI-24}}, pages 3602--3612. International Joint Conferences on Artificial Intelligence Organization.
\newblock Main Track.

\bibitem[{Xu et~al.(2024)Xu, Fei, Pan, Liu, Lee, and Hsu}]{xu-etal-2024-faithful}
Jundong Xu, Hao Fei, Liangming Pan, Qian Liu, Mong-Li Lee, and Wynne Hsu. 2024.
\newblock \href {https://doi.org/10.18653/v1/2024.acl-long.720} {Faithful logical reasoning via symbolic chain-of-thought}.
\newblock In \emph{Proceedings of the 62nd Annual Meeting of the Association for Computational Linguistics (Volume 1: Long Papers)}, pages 13326--13365, Bangkok, Thailand. Association for Computational Linguistics.

\end{thebibliography}

\appendix

\begin{table*}[t]
\centering
\tiny
\begin{tabular}{@{}p{1cm}p{4cm}p{7cm}p{1cm}p{1cm}@{}}  
\toprule
\textbf{Dataset} & \textbf{Problem} & \textbf{Explanation} & \textbf{Iteration} & \textbf{Validity} \\
\midrule
WorldTree & \textbf{Hypothesis}: Being able to read is an example of a learned trait. & Usually people learn how to read and write in school. trait is synonymous with characteristic. inherited characteristics are the opposite of learned characteristics; acquired characteristics. inheriting is when a inherited characteristic is copied; is passed from parent to offspring by genetics;DNA. &\centering 0\arraybackslash & Invalid \\\\

 & & Usually, people learn how to read and write in school, and activities learned in school are considered learned characteristics. Trait is synonymous with characteristic. Inherited characteristics are the opposite of learned characteristics. Learned characteristics are also known as acquired characteristics. Inheriting is when an inherited characteristic is copied or passed from parent to offspring by genetics or DNA. Learned characteristics are synonymous with learned traits. &\centering 2\arraybackslash & Valid \\
\midrule
Clinical & \textbf{Hypothesis}: BRCA2 is a human protein involved in double strand break DNA break repair via homologous recombination repair. & BRCA2 is a human protein involved in homologous recombination repair. Homologous recombination repair is a double strand break DNA repair process wherein damaged DNA is replaced by undamaged homologous molecules from sister chromatids or paternal/maternal copies of chromosomes. &\centering 0\arraybackslash & Invalid \\\\

 & & BRCA2 is a human protein involved in homologous recombination repair. Homologous recombination repair is a method used in double strand break DNA repair, wherein damaged DNA is replaced by undamaged homologous molecules from sister chromatids or paternal/maternal copies of chromosomes. BRCA2's involvement in homologous recombination repair directly contributes to double strand break DNA repair. &\centering 2\arraybackslash & Valid \\
\bottomrule
\end{tabular}
\caption{Examples of iterative explanation refinement for WorldTree and clinical explanations using GPT-4o and Isabelle.}
\label{tab:examples_appendix_table}
\end{table*}

\section{Explanation Refinement Examples}
\label{sec:appendix_explanation_refinement_examples}

Table \ref{tab:examples_appendix_table} shows additional examples of iterative refinement via hard critique (i.e. GPT-4o and Isabelle) on Worldtree and clinical explanations.

\section{Implementation Details}

\subsection{Data Model}
\label{sec:appendix_corpora}

\begin{figure}[t]
\centering  \includegraphics[width=0.9\linewidth]{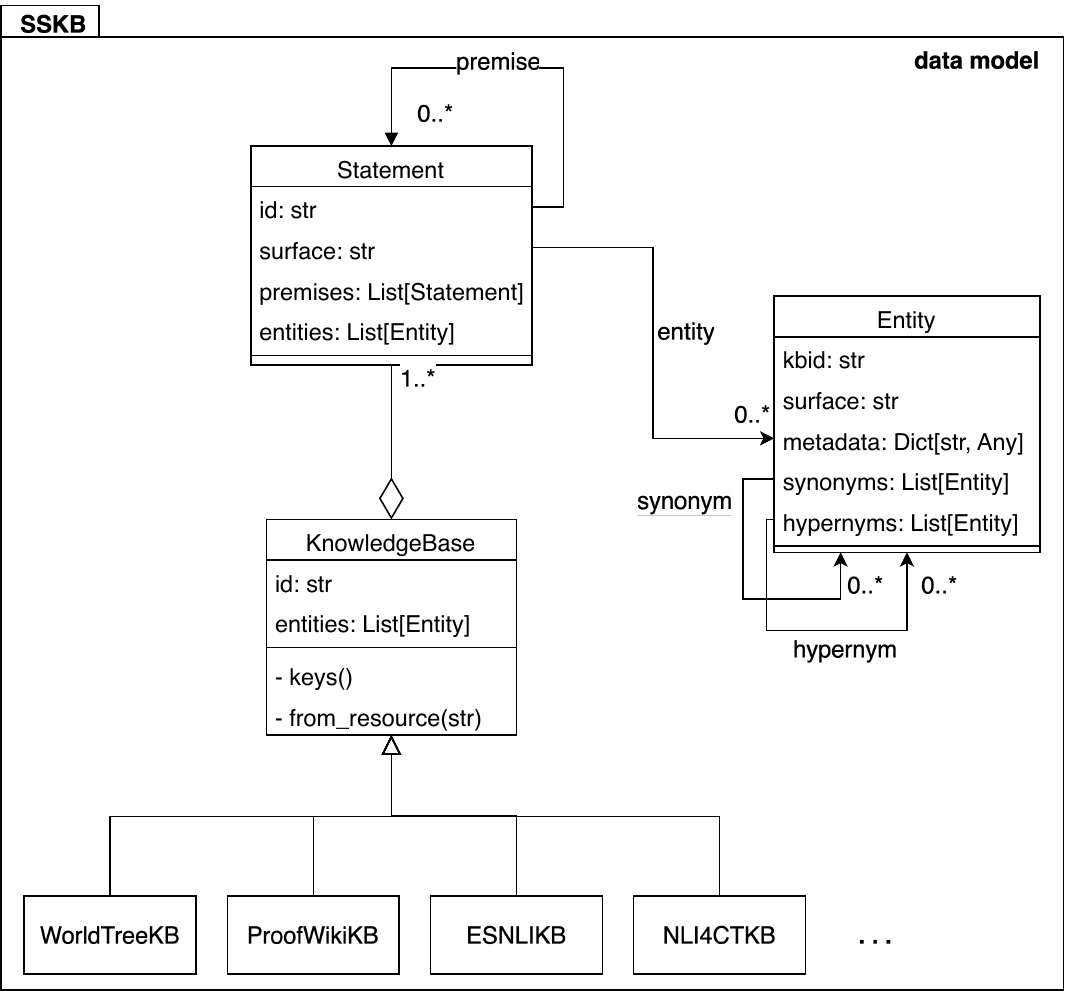}
  \caption{UML diagram of the \textit{Simple Statement Knowledge Bases} (SSKB) package. The classes at the bottom implement loading facilities for popular NLI datasets.}
  \label{fig:sskb_diag}
\end{figure}

The following code snippet shows an example of how to use SSKB to load data from external explanation corpora (i.e., WordlTree):

\begin{lstlisting}[language=Python]
from sskb import WorldTreeKB

kb = WorldTreeKB()

# Retrieve the individual facts in the corpus
facts_kb = [stt for stt in kb if (stt.annotations["type"] == "fact")]

# Retrieve the questions in the test set
test_questions = [stt for stt in kb if (stt.annotations["type"] == "question" and stt.annotations["split"] == "test")]

# Retrieve a complete explanation
explanation = [p.surface for p in test_questions[42].premises]
\end{lstlisting}

\subsection{Retrieval Models}
\label{sec:appendix_retrieval}

An example of how to instantiate and query the data model via BM25 is presented below:

\begin{lstlisting}[language=Python]
from retrieval.bm25 import BM25Model

# Initialize BM25 model
bm25 = BM25Model(facts_kb)

# Construct the list of queries
queries = [q.surface for q in test_questions] 

# Compute BM25 ranking and scores
res_bm25 = bm25.query(queries)
\end{lstlisting}
An example of how to instantiate and query the data model using an ensemble model is presented below:

\begin{lstlisting}[language=Python]
from retrieval.ensemble import EnsembleModel

# Initialise the ensemble model
ensemble_model = EnsembleModel(
        [bm25, unification],
        weights = [0.8, 0.2]
    )
\end{lstlisting}

\subsection{Generative Models}
\label{sec:appendix_generative}

An example of how to prompt GPT-4o for explanation generation is provided below:

\begin{lstlisting}[language=Python]
from generation.generative_model import GPT

# Parameters for prompting
api_key = "personal key"
prompt_file = "explanation_prompt.txt"

# Input problem
hypothesis = "I pricked the baloon."
conclusion = "The balloon expanded."

# Initialise the model
llm = GPT('gpt-4o', api_key)

# Generate an explanation 
explanation = llm.generate(
             prompt_file,
             hypothesis,
             conclusion
         )
\end{lstlisting}

An example of a dynamic prompt is provided below, with \texttt{hypothesis} and \texttt{conclusion} acting as variables that can be specified at runtime

\begin{lstlisting}
You are an expert on causal reasoning
and explanation. You will use causal
knowledge and commonsense to provide
logical explanations for the provided
causal reasoning scenarios.

For the hypothesis and conclusion
provided in the test example, let's
think step-by-step and generate an
explanation...

Test Example:

Hypothesis: {hypothesis}
Conclusion: {conclusion}
\end{lstlisting}

\subsection{Critique Models}
\label{sec:appendix_critique}

An example of how to instantiate a hard critique model via an external Isabelle solver and GPT-4o as formaliser is provided below:

\begin{lstlisting}[language=Python]
from critique.isabelle import IsabelleSolver

# Example from e-SNLI
premise = "A couple playing with a little boy on the beach."
hypothesis = "A couple are playing with a young child outside."
explanation = "little boy is a young child."

# Initialise the model
llm = GPT('gpt-4o', api_key)

# Initialise the critique model
isabelle = IsabelleSolver(
    generative_model = llm,
    isabelle_session = 'HOL'
    )

# Perform the critique
res = critique_model.critique(
    hypothesis, 
    premise, 
    explanation
    )
\end{lstlisting}

\subsection{Iterative Refinement}
\label{sec:appendix_refinement}

An example of how to instantiate a complete refinement process for 10 iterations is provided below:

\begin{lstlisting}[language=Python]
from refinement.refinement_model import RefinementModel

# Initialise the refinement process
refinement_model = RefinementModel(
    generative_model = llm,
    critique_model = isabelle
)

# Perform refinement for 10 iterations
res = refinement_model.refine(
    hypothesis = hypothesis,
    premise = premise,
    explanation = explanation,
    iterations = 10
    )
\end{lstlisting}

\end{document}